# Industry Led Use-Case Development for Human-Swarm Operations

Jediah R. Clark[1], Mohammad Naiseh[1], Joel Fischer[2], Marisé Galvez Trigo[2], Katie Parnell[1], Mario Brito[1], Adrian Bodenmann[1], Sarvapali D. Ramchurn[1], Mohammad Divband Soorati[1].

[1]University of Southampton, University Road, Highfield, Southampton, SO17 1BJ
J.R.Clark@soton.ac.uk; m.divband-soorati@soton.ac.uk
[2]University of Nottingham, Nottingham, NG7 2RD

**Abstract**

In the domain of unmanned vehicles, autonomous robotic swarms promise to deliver increased efficiency and collective autonomy. How these swarms will operate in the future, and what communication requirements and operational boundaries will arise are yet to be sufficiently defined. A workshop was conducted with 11 professional unmanned-vehicle operators and designers with the objective of identifying use-cases for developing and testing robotic swarms. Three scenarios were defined by experts and were then compiled to produce a single use case outlining the scenario, objectives, agents, communication requirements and stages of operation when collaborating with highly autonomous swarms. Our compiled use case is intended for researchers, designers, and manufacturers alike to test and tailor their design pipeline to accommodate some of the key issues in human-swarm interaction. Examples of an application include informing simulation development, forming the basis of further design workshops, and identifying trust issues that may arise between human operators and the swarm.

## Background

Recent developments in robotics have led to a new generation of robots with small sizes and costs. Currently, robotic systems can consist of large numbers of robots that collaborate autonomously to achieve a shared goal. These robotic systems, called swarms, have received attention from many domains, such as Robotics and Human-Computer Interaction and Law (Brambilla et al. 2013, Walker et al. 2016). These individual robots traditionally work together without any central control and act according to simple and local behaviour. Such cooperation between swarm robots leads to emergent behaviour, i.e., the collective behaviour of these robots, which enables them to solve complex tasks. Swarms hold the potential to solve many complex problems by connecting a group of aerial, ground or underwater robots such as surveillance and payload delivery (Schranz et al., 2020). These systems promise solutions for missions in which direct human involvement could introduce significant risk and harm, e.g., search-and-rescue, firefighting, planetary exploration, and ocean restoration. Although swarms can perform their tasks autonomously, human involvement is still a requirement for implementing swarms in real-world scenarios. In other words, swarm operators are still responsible for monitoring and controlling the swarm to manage the complexity of a task that might be beyond the swarm's capability or for potential errors and failures of the swarm. Further, some application domains, such as healthcare and defence, require human supervision due to legal and ethical concerns (Verbruggen, 2019).

The distributed and collaborative nature of the swarms has led to several advantages, such as their adaptability to environmental changes, robustness against a single point of failure and scalability (Saffre et al., 2021). However, these advantages have presented novel challenges for effective Human-Swarm Interaction (HSI) that are only beginning to be addressed (Saffre et al., 2021). For example, when operators are tasked with operating a swarm of 50-100 robotic agents, issues arise such as how the swarm can remain strategically relevant, without violating operational boundaries and remaining in a working condition. Underpinning these issues is the fundamental basis of how operators calibrate and maintain trust with the system (Nam et al., 2018). Recently the HSI community has highlighted the need for consistent language and definitions; clearly defined operators' roles, responsibilities, and HSI evaluation metrics. For instance, Harriott et al., (2014) pointed out the lack of rigour in evaluating the quality of HSI. The study also presented HSI evaluation metrics derived from biological swarm literature. Further, Kolling et al. (2015) presented several considerations to be made when developing Human-Swarm interfaces. Their research discussed several use-cases of Human-Swarm Interaction from the perspective of remote human supervisors through reviewing existing academic literature. While these generalized frameworks are steps in the right direction, two key gaps remain undefined in most existing work on Human-Swarm Interaction:

- HSI studies are often from a "general-purpose" perspective with a broadly defined goal of interaction, not to address specific needs of real-world use-cases.
- HSI studies are not evaluated to adequately reflect how effective their approaches are in real-world settings. Barring a few exceptions (Patel et al., 2019, Bjurling et al., 2020), much of the existing work is designed and developed for benchmark and validated with user studies limited to users in research settings such as Amazon Mechanical Turk, e.g., (Liu et al., 2019).

The current stage of the literature is a body of methodological work, e.g., control and interaction techniques, without clear use-cases and established real-world utility. In this paper, we argue that real-world case studies from current industrial organizations are needed to inform the efficacy of Human-Swarm Interaction research. This study aims to explore real-world use-cases from industrial organizations of Human-Swarm partnership where humans are responsible for controlling and monitoring the swarm. We seek to define operational boundaries, operators' roles and their implications for Human-Swarm trust. We argue that such knowledge is essential to understand how global organizations intend to make use of robotic swarms, and to what extent they foresee barriers in trust and operation. To that end, this research study seeks to:

- Gain insight into the needs and requirements of global organizations working with swarm robotics.
- Provide the community with industry-developed scenarios that are suitable for design and evaluation tasks.
- Identify how a user may interact and develop trust with the swarm, what communication links will be required, and how a robotic-swarm system may operate within real-world situations.
- Discuss research gaps by comparing the existing body of work to the need of the use-cases.
- Propose research directions to develop effective Human-Swarm Interfaces that would lead to improved Human-Swarm performance and consequently improved trust scores.

The use-cases generated in this research study aim to provide researchers, practitioners, and manufacturers with a basis in which autonomous robotic swarms can be designed and regulated within future operational scenarios. The specified tasks and activities identified within each use case are collated to provide an accessible reference tool for the deployment of robotic swarms. Our use cases can also be used as a probe in Human-Swarm studies to explore new areas of research that remain undefined in Human-Swarm settings. For instance, explainability has been previously identified as the main factor that affects Human-Robot trust. However, it remains unclear if the explanations studied in Human-Robot settings would be applicable to Human-Swarm teaming. More fundamentally, what kind of questions an explainable swarm should be expected to answer, and what those explanations would look like, are currently unknown. To that end, our generated use-cases may be used to feed into scenario-based elicitation methods (Holbrook, 1990) to explore how a swarm of robotics can explain their behaviour to human operators in dynamic environments.

The paper is structured as follows. Section 2 describes the research method, including the sample, material, and instruments used in our study. In Section 3, we present our results and findings. Finally, in Section 4, we discuss the implications of the findings on future research and development in the field and conclude the paper.

## Method

Ethical approval was acquired via the Ethics Committee of the School of Computer Science from the University of Nottingham [Reference No. CS-2020-R54].

### Participants

Participants were recruited by advertising to the project partners of the Trustworthy Autonomous Systems Hub representing various expertise within the industry. 11 participants took part in the workshop (10M, 1F), comprised of professional experts in the research, and development, operation, and management of unmanned vehicle systems. Experts were members of either defence or maritime organizations.

### Design

The workshop was organized as an online event, with participants joining an introductory presentation by the research team and then being split into three break-out rooms to work collaboratively on three tasks. Within each breakout room participants collaborated on 'Miro boards', an online visual interface for collaborating on shared projects (Miro, 2022). Participants in each group were tasked with generating their own use-case that could include homogenous or heterogenous vehicle systems consisting of aerial, land, and maritime vehicles. Swarms were delineated from multi-Unmanned Aerial Vehicle (UAV) systems by outlining that swarm robotics include autonomous behaviours of multiple agents (e.g., flocking, autonomous search patterns, auto-allocation of tasks), rather than individually being controlled. The construction of scenarios was unrestricted, as long as a task involving at least one human operator and a swarm of Unmanned Vehicles (UVs) that were not able to be individually controlled to ensure task success was defined. Participants were encouraged to use their professional experience to guide the construction of use-cases. Break-out groups were tasked with addressing three core themes: the scenario, operational boundaries, and implications for trust, as defined in Table 1.

### Procedure

The workshop lasted three-hours in total. Participants were given a ten-minute introduction to inspiration material consisting of links to news articles related to swarms and a table outlining an introduction to the various domains in which swarm robotic behaviors are currently in-place (e.g., Schranz et al., 2020). Once introduced to the workshop and the agenda, participants were split into three separate groups consisting of 3-4 members. Each group took part in three activities for use-case co-creation involving: scenario formation, operational boundaries, and identifying trust factors when interacting with the swarm in their identified scenario. After each section, each group presented their work on each activity back to all attendees of the workshop. Presentations after each session were audio recorded to compile use-cases.

Table 1. Group Defined Scenarios and Tasks

| | |
|---|---|
| Scenario Generation | *"Outline the scenario or problem your swarm will be tasked with. Specify the swarm-based solution, the people involved, and an overview of the system (i.e., how it works, what information is being processed, how does it address the task)"* |
| Operational Boundaries | *"Identify the limits, boundaries and capabilities of the system including the physical and cognitive arrangements of the system (e.g., roles, objects), how communication will occur, and what limitations and boundaries are present (e.g., range, missing information, dropouts, workload)"* |
| Implications for Trust | *"Define the main factors that may affect trust during the interaction with the swarm in the use case. Please the previous tasks"* |

**Analysis**

Miro-Boards were compiled into task diagrams split across three time periods: planning, exploring and response. All references to actions, communication and requests were included to populate the diagrams, separated by agents (e.g., swarm, supervisor, analyst, first responders). Use cases were then compiled to define a single scenario that can be applied to simulation platforms. We followed an iterative process across several research meetings to formulate, combine, and conceptualize the emerged concepts. This iterative process was put in place to examine and ensure that the final use case was representative of a possible real-life scenario.

# Results

**Group Scenarios**

Each break-out group outlined mission descriptions, timelines, team-roles, and tasks for their scenario. Each use case focused on casualty response tasks resulting from a disaster; two of which defined a search-and-rescue task, whereas one focused on the delivery of medical equipment to appropriate locations. Table 2 outlines each group's defined scenario and task definition.

*Compiled Use-Case Background*

"A hurricane has affected a settlement. Many were evacuated, but some did not manage to escape and have been caught in debris and fallen structures. A UAV swarm is being deployed to identify the locations of survivors, evaluate their condition, and advise on where unmanned ground rescue vehicles should be deployed to evacuate casualties."

*Compiled Use-Case Agents and Tasks*

Figure 1 outlines the sequence of tasks required to ensure task success, drawn from each group's use-case definitions. The figure presents four agents responsible for mission success, represented by four vertical orange bars within the figure:

- The operator – responsible for the operational performance of the swarm including the allocation of UAVs to airspace, allocation of tasks, and monitoring the connectivity and battery status of the swarm.
- The swarm – responsible for locating casualties and relaying data about the condition of the casualty and the environment. It will communicate information related to uncertainty, object classifications, and images of the scene.
- The analyst – tasked with sifting through incoming data from the swarm and screening for accuracy. Hazardous objects, route blockages, and the location and criticality of the casualty (calculated by movement, heat signatures etc.) are presented to the analyst during the mission. The analyst will then identify suitable locations for first-responders or unmanned ground vehicles (UGVs) to rescue high-priority casualties.
- The responders/UGVs – will take on the commands or recommendations (if a human responder) of the analyst to arrive at the target location and evacuate casualties.

Table 2. Group Defined Scenarios and Tasks

| Group | Scenario Description | Operators' Tasks |
|---|---|---|
| 1 | A multi-human team consisting of an operator (tasked with controlling the swarm) and an analyst (tasked with analysing processed data) seek to identify civilians and threats in a reconnaissance mission. | Allocate swarm, receive object classifications, query object classifications, send mission-information. |
| 2 | Operators work with coordinators and the swarm to evaluate target areas and suitable landing zones for medical supplies, whilst ensuring that airspace and other parties are managed. | Allocate swarm, search for target zone, identify suitable drop-off zone, call-in delivery drone, deliver payload. |
| 3 | Locating and evacuating casualties. Identify, classify, and prioritize casualties based on criticality. Deploy ground vehicles to appropriate locations. | Allocate swarm, relay conditions, classify casualties, plan rescue route, and deploy rescue vehicle. |

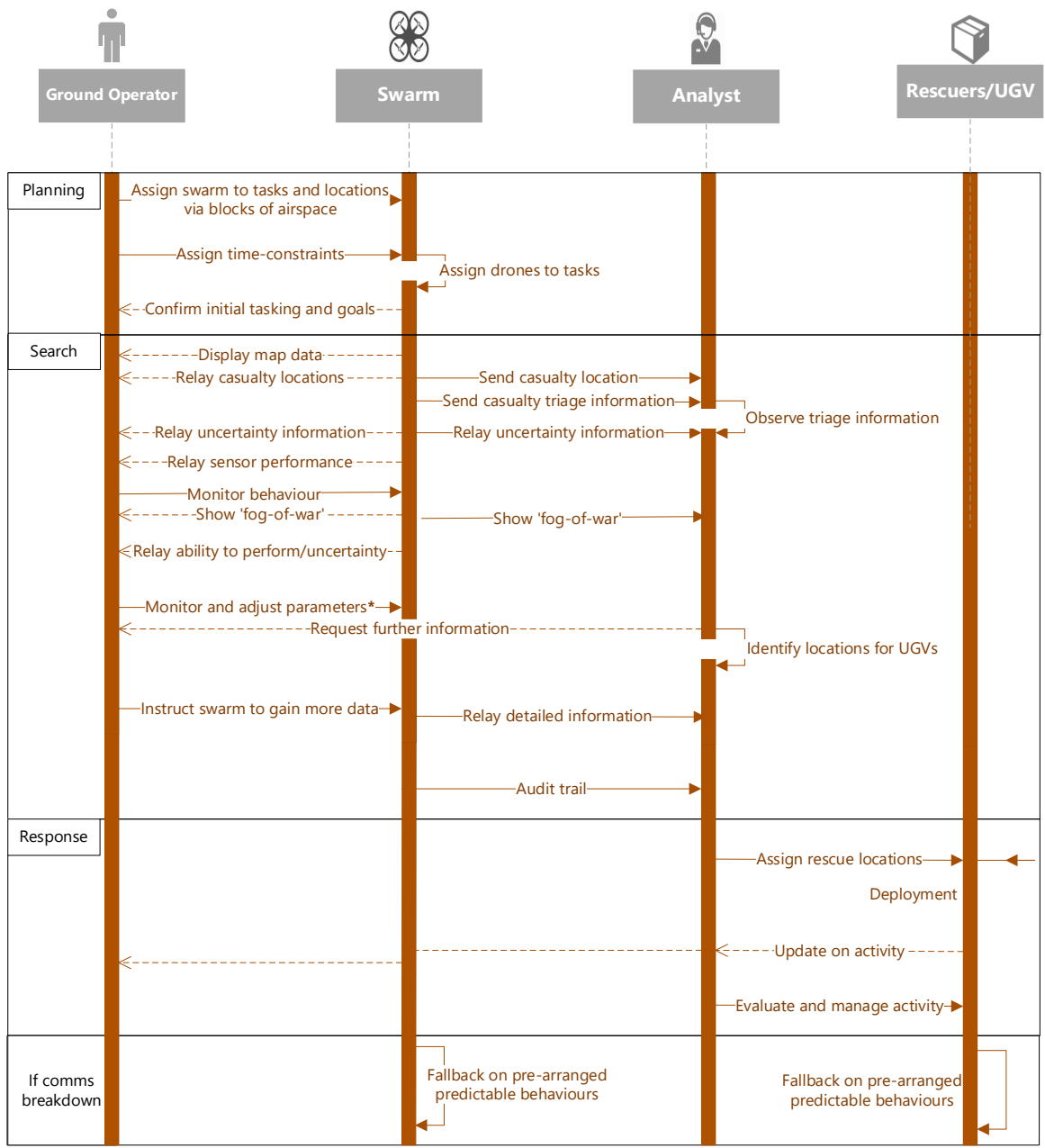

Figure 1. Task-flow diagram for the final collated use-case

The figure is laid out into three stages, which are populated with tasks that relay to another agent within the diagram, representing the flow of information and activities.

- The Planning Stage – The period when the operator and analyst familiarize themselves with the scene and any information made available to them before the search. The operator allocates the swarm to their search locations and gives groups of UAVs tasks to complete. The swarm relays that they have received their tasks. This can be done via a map to decide which areas may be more likely to have people (schools, housing blocks, leisure sites).
- The Search Stage – The main component of the use case. The swarm enacts the search for casualties whilst relaying casualty positions and images of the scene. Casualty information is relayed to the analyst for them to triage casualties into rescue-priority. Activity information such as the task being conducted, the location of UAVs, and future intentions are relayed to the operator. Additional information such as uncertainty (of object classification or location of objects) and operational state (such as the battery, connectivity, and structural integrity) are relayed so that the operator can monitor and adjust parameters accordingly. These parameters might include battery usage, swarm density to increase connectivity, or the operator might simply recall UAVs that are low in battery back to the base-station to reduce the frequency of dropouts. Both the operator and analyst can request additional information regarding the objects in the scene, or request images to be taken for additional clarity. The operator may adapt the actions of the swarm in-line with the analyst's needs so that detailed information can reach the analyst when making decisions on where to send rescue vehicles. During this process, an audit trail is logged so that the analyst can keep track of swarm activities.
- The Response Stage – Once the analyst is satisfied with the incoming data, they can allocate rescue locations for either first-responders or ground vehicles to evacuate as many critical casualties as possible. This decision is to be made by selecting locations that are accessible and can optimize the distance between itself and casualties.

Should communication deteriorate within the swarm, behaviours will fall back on pre-defined instructions set by the operator (e.g., follow the last instruction and patrol the area in which it no longer has an instruction until connectivity is restored).

## Discussion

Despite the existence of HSI studies investigating human-swarm trust, their efficacy in improving real-world use cases is yet to be sufficiently explored. The use-case scenarios and the compiled use-case in this research study are the first steps toward that goal by providing researchers, practitioners, and manufacturers with a basis in which autonomous robotic swarms can be designed and regulated within future operational scenarios. This use case is driven by professional UAV operators and designers to reflect the current trends in unmanned-vehicle technological development. Communication requirements and distributed role allocation are provided to map-out the potential for multiple humans to work with a swarm, and outlines what a future human-swarm mission might entail.

It is critical that the HSI domain develops past that of human-robot interaction, as much of the research in human-robot explainability (Naiseh et al. 2021) focuses on individual robot agents which may not be directly applicable to HSI. For example, swarm display methods have an increased requirement for the concatenation of information and succinct display methods to reduce workload and optimize trust and usability (Divband Soorati et al., 2021; Saffre et al., 2021). Explainability in swarms can be used to better calibrate trust by giving the operator and analyst a better understanding of the actions and decisions of the swarm. Our integrated use case can be used as a baseline scenario in human-swarm studies to explore new areas of research that remain undefined in human-swarm settings.

The compiled use-case is presented to provide representation of a real-world scenario with a defined storyline, and phases of the mission. This can be used to test multiple research questions that are facing the HSI community as autonomy increases. For example:

- How can the distributed awareness and cognition (in-line with modern theories of situation awareness; Stanton et al., 2017) of multiple human agents be specified and applied?
- What uncertainty information is required and how can this be communicated to human agents?
- How should interfaces and interactions be designed to support optimal trust, workload, and awareness?
- What procedures will be required when managing the swarm and processing outcoming information?
- How can human-swarm explainability be used to allow the operator to efficiently monitor the status and activities of the swarm?
- What fallback behaviours are most appropriate for managing the loss of connectivity?

Many of these questions can be addressed by rolling out modern testbeds for HSI to be tested in real-time with multiple users taking part in the same simulated mission. Additionally, each phase of the mission can be addressed separately when managing human-swarm interactions. For example, the planning phase can be explored to develop efficient communication methods and task allocation, and how supporting information (e.g., map displays) can be used for supporting planning and coordination activities. Individual components of the use-case can also be drawn to test hypotheses, for example, focusing on what information will need to be shared or exclusive to either an operator or an analyst (e.g., should an analyst be aware of how the swarm is performing, or should an operator know what casualties are being investigated by the analyst?). Additionally, communication modalities should be considered, for example, how would the analyst and the operator communicate to manage the task and swarm activities, and how can this be facilitated through interfacing (e.g., text, voice, or in-interface communication). Additionally, considering how objects are classified and relayed to the human-operators should be considered (e.g., how many swarm members agree or disagree on the object classification or incoming information?).

The use-case serves as a scenario to test these pressing questions and allows multiple communication links and performance metrics to be assessed. In this scenario, the speed at which the team locate casualties, identify criticality and the selection of an optimal pick-up location can all be factored in when analysing performance. Additionally, researchers can analyse and monitor communication between each agent to understand what and when information is communicated. The use-case can be applicable to civil applications (e.g., environment monitoring, search and rescue, surveillance) as well as defence applications. To that end, the use case can be flexibly applied to many working domains.

## Acknowledgements

This work was conducted as part of the Trustworthy Autonomous Systems Hub, supported by the Engineering and Physical Sciences Research Council (EP/V00784X/1).